# Automated Diagnosis of Lymphoma with Digital Pathology Images Using Deep Learning

Hanadi El Achi, MD[1]; Tatiana Belousova, MD[1], Lei Chen, MD[1]; Amer Wahed, MD[1];
Iris Wang, MD[1]; Zhihong Hu, MD[1]; Zeyad Kanaan, MD[2], Adan Rios, MD[2],
Andy N.D. Nguyen, MD[1*]

[1]Department of Pathology and Laboratory Medicine, University of Texas Health Science Center at Houston, Texas, TX 77030
[2]Department of Oncology, University of Texas Health Science Center at Houston, Texas, TX 77030

## Abstract

Recent studies have shown promising results in using Deep Learning to detect malignancy in whole slide imaging, however they were limited to just predicting positive or negative finding for a specific neoplasm. We attempted to use Deep Learning with a convolutional neural network (CNN) algorithm to build a lymphoma diagnostic model for four diagnostic categories: (1) benign lymph node, (2) diffuse large B-cell lymphoma, (3) Burkitt lymphoma, and (4) small lymphocytic lymphoma. Our software was written in Python language. We obtained digital whole-slide images of Hematoxylin and Eosin stained slides of 128 cases including 32 cases for each diagnostic category. Four sets of 5 representative images, 40x40 pixels in dimension, were taken for each case. A total of 2,560 images were obtained from which 1,856 were used for training, 464 for validation, and 240 for testing. For each test set of 5 images, the predicted diagnosis was combined from prediction of five images. The test results showed excellent diagnostic accuracy at 95% for image-by-image prediction and at 100% for set-by-set prediction. This preliminary study provided a proof of concept for incorporating automated lymphoma diagnostic screen into future pathology workflow to augment the pathologists' productivity.

## INTRODUCTION

Lymphoma is a clonal malignancy of lymphocytes, either T cells or B cells. The different lymphoma entities are typically first suspected by their pattern of growth and the cytologic features of the abnormal cells via light microscopy of Hematoxylin and Eosin stained tissue sections. Immunophenotyping is typically required for diagnosis with flow cytometry and/or immunohistochemical stains. In addition, cytogenetics, molecular pathology results, and clinical features are often needed in finalizing the diagnosis in certain lymphoma types [1]. Lymphoid malignancies are diagnosed in 280,000 people annually worldwide and include at least 38 entities according to the World Health Organization (WHO) Classification of Lymphoid Malignancies [1]. Due to subtle differences in histologic findings between various types of lymphomas, histopathologic screen often presents a challenge to the practicing pathologists.

*Corresponding author
Andy N.D. Nguyen, MD
Department of Pathology and Laboratory Medicine
University of Texas Health Science Center at Houston
6431 Fannin Street, MSB 2.292, Houston, TX, 77030
(713) 500-5337; fax (713) 500-0712
Nghia.D.Nguyen@uth.tmc.edu

The recent introduction of digital Whole Slide Imaging (WSI) opens an opportunity for automated identification of histopathologic features of lymphomas [2]. The WSI systems digitize whole glass slides with stained tissue sections at high resolution, helping pathologists in microscopic examination [3]. The quality of the images is pivotal for optimal microscopic interpretation. Fortunately, digital image acquisition has improved substantially in recent years with the implementation of instrumentation capable of acquiring data at very high rates and with excellent resolution [2]. Recently, the Food and Drug Administration (FDA) cleared the marketing of the first WSI system for digital pathology diagnosis beyond the scope of research [4]. The image interpretation process of digital slides is being actively studied in diagnostic medicine particularly with the advent of machine learning techniques which made considerable contributions to the realm of Pathology and the birth of a novel and highly sophisticated field known as "Digital Pathology" (DP). One of the definitions of DP is a process of converting histology glass slides to digital slide images with high-resolution using whole slide scanners, followed by interpretation and generation of pathologic information by machine learning techniques [3].

Machine learning, a branch of artificial intelligence (AI), provides automated methods for data analysis. The principle of the technique is based on the ability of the machine to learn information from previously saved data in databases and improve itself for better diagnostic findings [5]. Machine learning frameworks have considerably evolved throughout the decades; the first conventional AI algorithms included Support Vector Machine (SVM) and Neural Network (NN). These techniques were followed by the new sophisticated Deep Learning (DL) algorithms such as Convolutional Neural Network (CNN), Recurrent Neural Network (RNN), Long Short Term Memory (LSTM), and Extreme Learning Model (ELM) [5]. DL, the newest subgroup of machine learning, has largely demonstrated itself as the most effective and reliable machine learning technique when applied to the medical field. It is a growing innovation trend in data analysis that has been termed one of the ten breakthrough technologies of 2013 [6]. Since DL presents in many algorithmic formats, it cannot be considered a single technique. Instead, DL can roughly be described as the latest generation of artificial neural networks, consisting of specially designed layers that permit higher levels of abstraction and improved predictions from data input [7]. DL is based on the principle of neural network with the neuron being the fundamental unit. This neural network forms the so-called "layered architecture" made up of multiple layers of neurons, recently reaching 1000, lying between input and output layers. Each neuron receives the input data from multiple neurons of the previous layer and then uses unsupervised learning to find certain characteristic features that will be filtered and added together to ultimately generate an output to be communicated to the next layer. Increasing the number of layers allows for more features to be detected, and more complex patterns to be learned [8]. DL has been applied to a wide range of domains, from speech recognition [9-13] to image analysis [14-16], and natural language processing [17-19]. In recent years, DL techniques have become the state of the art in computer vision. A specific DL neural network subtype, the convolutional neural networks or CNN [20-21], has become the de-facto standard in image recognition and has been shown to approach human performance in various tasks [7]. These CNN systems excel by learning relevant features directly from raw data in large image databases; this contrasts with the more traditional pattern recognition techniques, which rely on detecting manually-crafted quantitative features [3]. Recent studies showed that the generic descriptors extracted from CNNs are extremely effective in object recognition and localization in digital images. Medical image analysis groups around the world started to apply CNNs and other DL methodologies to a wide range of applications [14-16], and promising results have been emerging from recent studies [7, 20-23]. The International Symposium on Biomedical Imaging (ISBI) held the Camelyon Grand Challenge [22] in 2016 to evaluate computational systems for the automated detection of metastatic breast cancer in WSI of sentinel lymph node biopsies. The Harvard & MIT team won the grand challenge obtaining an Area Under the receiver operating Curve (AUC) of 0.925 for the task of WSI classification, i.e. positive versus negative for metastasis for each slide. In a Stanford University study using DL network to diagnose skin cancers [23], the research group used biopsy-proven clinical images to successfully diagnose two critical binary classifications of skin cancers, (a) keratinocyte carcinomas versus benign seborrheic keratoses showing an AUC of 0.96, and (b) malignant melanomas versus benign nevi with an AUC of 0.94. Minot et al. in their study of an automated cellular imaging system for assessing HER2 status in breast cancer specimens showed that automated image analysis provides a higher concordance rate with FISH than visual inspection for breast cancer [24]. Other studies involved prognostication and Gleason scoring [25] for prostate cancer, and the assessment of the Ki-67 labeling index [26] for meningiomas showed promising results. Two studies reported successful interpretation of Human Epidermal Growth Factor Receptor 2 (HER2) via automated image analysis [27-28].

Hematopathology has also earned its part in the digitalization movement. Recent projects have shown promising results using machine learning to detect lymphoma with WSI. However, studies involving the application of DP for lymphoma detection are still limited to just positivity versus negativity for a particular neoplasm [29-30]. In this study, we explore how DL can be used to accurately classify a test case as one of four lymphoid entities representative of various morphologic patterns in lymphoma: benign lymph node, diffuse large B-cell lymphoma (DLBCL), Burkitt lymphoma (BL), and small lymphocytic lymphoma (SLL).

## MATERIALS AND METHODS

We obtained WSIs from two data sources including Virtual Pathology at the University of Leeds [31] which contains 355,966 WSI collections (114.92 TB of data), and Virtual Slide Box from University of Iowa [32], with over 1,000 WSI collections hosted by MicroBrightField Bioscience (Williston, VT USA) on Biolucida Cloud Portal. The WSIs on both websites were obtained with Aperio WSI systems (Aperio Technologies, San Diego, CA, USA). For the Virtual Pathology collection at the University of Leeds, we used Chrome web browser to view the images at 40x magnification, and used SnagIt software (TechSmith Corp, Okemos, Michigan, USA) to capture 40x40 pixel image patches at random locations on the histologic section. Similarly, for the Virtual Slide Box collection at Iowa University, we viewed the images at 40x magnification with the Biolucida viewer and used SnagIt software to capture 40x40 pixel image patches at random locations on the histologic section. Each image patch is represented as a 40x40 matrix (40 rows and 40 columns) representing the intensity of 1,600 pixels. The image file was subsequently converted into a one-dimensional file with 1,601 entries; the first entry in the file stores the diagnostic label of the image, and entries from 2 to 1,601 store all the pixel intensity values. Our study included 32 cases for each of the following entities: benign lymph node, DLBCL, BL lymphoma, and SLL. Representative WSI sections for these diagnostic categories are shown in Figure 1. The diagnostic labels were as following: 0 for benign lymph node, 1 for DLBCL, 2 for BL lymphoma, and 3 for SLL. A total of 128 cases were used in this study. Four sets of 5 representative 40x40 pixels images were captured for each of the 128 cases, giving a total of 2,560 images.

CNN systems for image recognition have greatly benefited from the use of parallel processing because most computations for image operations are based on matrix operations [33]. Parallel processing can significantly decrease computing time by performing all similar matrix operations at the same time instead of in sequence. The computer graphics cards, known as Graphics Processing Units (GPUs), contain hundreds or thousands of processing cores and bring great increase in computational speed. We designed a CNN model in Python language [34], an object-oriented programming language most commonly used in deep learning. We also used TensorFlow [35] and Keras [36], two important Python libraries particularly useful in DL modelling. TensorFlow, used as backend for our software, allows for parallel computing using GPU. Our computing platform included a Personal Computer (PC) with Intel i5-4590, 8GB RAM, Microsoft Windows 8-64 bit. The PC's GPU is a GTX745 (4 GB), an NVIDIA card with 384 cores supported by Compute Unified Device Architecture (CUDA) [37]. The core element of the CNN algorithm is convolution [38], an operation in image processing using kernels (filters), to detect or modify certain characteristics of an image including options such as "smooth", "sharpen", "intensify", or "enhance". Mathematically, a convolution is done by multiplying the pixels' value in the image patch by a kernel matrix; this effectively enhances the value of an image patch by adding the weighted values of all the neighboring pixels together. By moving the kernel across input image, one obtains the feature map as a filtered image. As shown in Figure 2, the CNN model [39] has the following processing pipeline for the detection of visual categories: the convolutional layers perform feature extraction consecutively from the image patch to higher level features, followed by the max-pooling layers' down-sampling to reduce the amount of computation in the network, finally the last fully-connected layers provide prediction based on the given features. Nodes in the fully connected layers have connections to all activations in the previous layer, as seen in traditional neural networks.

For our CNN network, 1,856 images out of 2,560 were used for training the model. 464 images were used for validation, and the remaining 240 images were used for testing. For each test set of 5 images, the predicted diagnosis was combined from the prediction of all five images, i.e. at least three or more must agree to be considered as the predicted result, a process known as "majority voting" [40].

**Fig. 1** Representative WSI sections for four diagnostic categories in our deep learning system

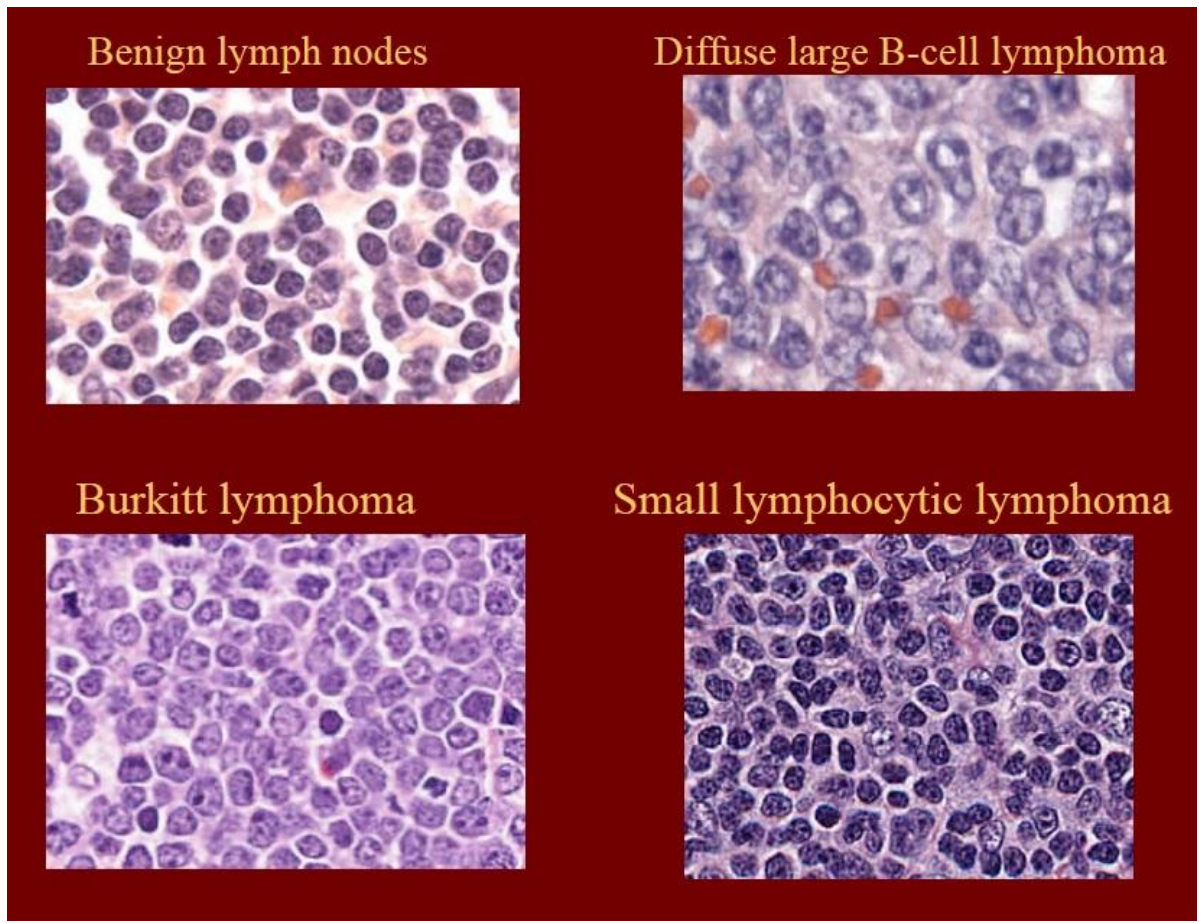

**Fig. 2** Processing pipeline of a convolutional neural network for the detection of visual categories in images

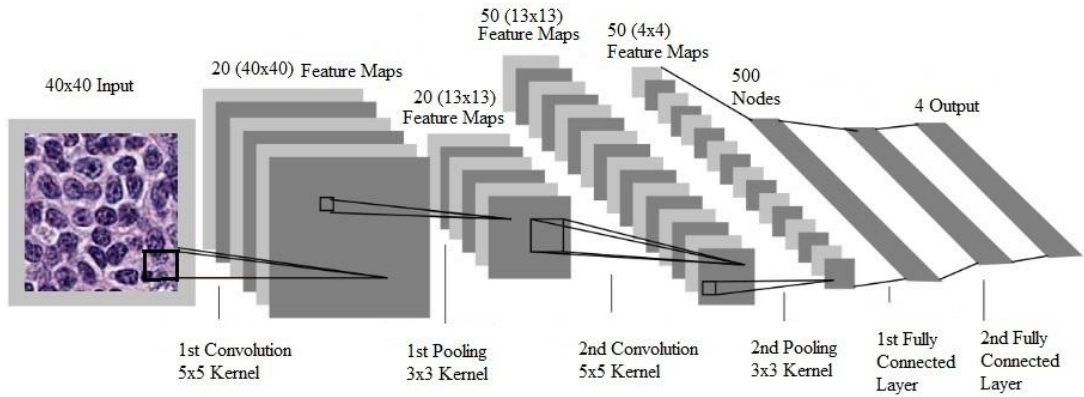

## RESULTS

Image-by-image scoring was first performed by selecting single random images among the selected cases for testing. Out of 240 test images, a total of 228 images were correctly diagnosed by the CNN model, and the remaining 12 images were given incorrect diagnosis, yielding an overall 95% accuracy for diagnostic prediction (Table 1). Among the 12 images with lack of concordance between the observed and the predicted diagnosis: 4 SLL images were predicted as benign, 4 other SLL images were predicted as DLBCL, and 4 benign images were predicted as BL. Set-by-set scoring performed by incorporating all five images for every set and implementing the majority voting strategy -at least three out of the five representative images of the set must agree- yielded an accuracy of 100% with 48 sets out of 48 being correctly diagnosed (Table 2). It appears that diagnosis based solely on one image is too stringent to be of practical value. Instead, the microscopic diagnosis needs to be based on all the five representative images to exclude outliers for a given set.

The optimization of CNN algorithms requires careful fine-tuning of network configuration and learning parameters (called hyper-parameters); this fine-tuning typically necessitates trial-and-error, and sometimes brute-force search [41]. During network training in this study, we have attempted various configurations for the CNN network to achieve optimal features and accuracy. As illustrated in Figure 2, we noted that our DL network performed best with an optimal set of hyper-parameters for the CNN layers [3,33] as following:
-1st convolutional layer: 5x5 kernel, 20 feature maps, activation function: tanh
-1st pooling layer: 3x3 kernel, 3x3 stride, pooling function: max-pooling

-2nd convolutional layer: 5x5 kernel, 50 feature maps, activation function: tanh
-2nd pooling layer: 3x3 kernel, 3x3 stride, pooling function: max-pooling
-1st fully connected layer: hidden nodes: 500, activation function: tanh
-2nd fully connected layer (output layer): 4 nodes, activation function: softmax

Table 1. Accuracy in predicting diagnosis using one single image at a time

| Observed Diagnosis | | | | | |
|---|---|---|---|---|---|
| **Predicted Diagnosis** | | **Benign** | **DLBCL** | **BL** | **SLL** |
| | **Benign** | **56** | 0 | 0 | 4 |
| | **DLBCL** | 0 | **60** | 0 | 4 |
| | **BL** | 4 | 0 | **60** | 0 |
| | **SLL** | 0 | 0 | 0 | **52** |

Accuracy: 228/240=95%

Legends:
DLBCL: diffuse large B cell lymphoma
BL: Burkitt lymphoma
SLL: small lymphocytic lymphoma

Table 2. Accuracy in predicting diagnosis for sets of 5 images using majority voting
(3 out of 5 images for each set must agree)

| | | Benign | DLBCL | BL | SLL |
|---|---|---|---|---|---|
| **Observed Diagnosis** | | | | | |
| **Predicted Diagnosis** | **Benign** | 12 | | | |
| | **DLBCL** | | 12 | | |
| | **BL** | | | 12 | |
| | **SLL** | | | | 12 |

Accuracy: 48/48=100%

Legends:
DLBCL: diffuse large B cell lymphoma
BL: Burkitt lymphoma
SLL: small lymphocytic lymphoma

**DISCUSSION**

Lymphomas are a heterogeneous group of malignancies that account for 3.37% of all malignancies worldwide [42]. They are grouped into two large entities, Hodgkin lymphomas and non-Hodgkin lymphomas (NHL). Advances in technology revealed multiple subtypes of NHL including DLBCL that accounts for the largest subtype, follicular lymphomas as the second most common, BL and SLL as the relatively common subtype [42]. Due to subtle difference in histologic findings and difficulties for human eye distinguishing between various types of lymphomas, histopathologic screen often presents an arduous task to the pathologists and is susceptible to inter-observer and inter-laboratory variability [43]. Moreover, lymph node diseases are not restricted to malignancies, reactive and inflammatory changes due to infections which can have similar clinical and pathological presentation as lymphomas should always be part of the differentials. Thus, there is an urgent need to relieve the workload on pathologists by sorting out benign cases and giving them more time to focus on the more challenging tasks.

An automated diagnostic system for digital hematopathology images would be helpful to assist the pathologists in daily work. Previous attempts to classify histologic images were based on specific criteria such as nuclear shape, nuclear size, texture, etc. obtained by edge detection, and cell segmentation [3]. However, they were not very successful; attention has shifted to machine learning and specifically DL. DL neural networks for image recognition have recently gained significant research interest due to the development of CNNs and the advent of efficient parallel processing by modern GPUs. The core element of a CNN lies in its deep layers, which allow for extracting a set of discriminating features at multiple levels of abstraction [8]. Although DL is an active research field, its application to microscopic diagnosis of tumors is relatively new. Most published work has focused on diagnosis between two disease entities, or between benign tissue and one specific tumor, making it difficult to assess the practical value of the designed CNNs. The hematopathology part of the digitalization movement was limited overall to the sub-classification, and grading of lymphomas. Fauzi et al. [29] conducted a project for the grading of follicular lymphoma with the aid of computerized systems and confirmed the usefulness of the method in tissue grading. Another study using the Aperio AT2 instrument for WSI scanning with a newly developed algorithm for image analysis showed 82.5% concordance between the pathologists and the trained algorithms for subtyping of DLBCL [30]. To the best of our knowledge, only one robust study was conducted by Nikita et al. to classify lymphomas in one of the following three types: SLL, follicular lymphoma, and mantle cell lymphoma using spectral analysis with weighted-neighbor distance (WND) algorithm [44]. This study reported a high accuracy rate of 99%. However only a small number of 30 lymphoma cases were used which did not provide an adequately vigorous validation for the model. Our project was the first to get closer to actual practice by exploring how DL can be used to accurately classify a test case as one of the four representative entities of various morphologic patterns in lymphoma. We also include a substantial number of cases (128) and images (2,560). Our DL network with CNN algorithm yielded an impressive result with an accuracy of 100% when 12 sets of five images for every diagnosis were analyzed. We noted that selecting one single image for diagnostic prediction did not always show successful results (the accuracy was only at 95%). This finding emphasizes the subtle differences between the various types of lymphomas; they reflect the importance of whole slide scanning for a better prediction of the diseases, i.e. including images of many random fields of the slides to reach an accurate diagnosis. Since generic machine learning algorithm [39] is a key element with the CNN method, there is no need for manual settings of morphologic parameters for a specific tumor type (histologic pattern, nuclear architecture, shape, and texture, etc.). Subsequently the results from this study can be applied to other histopathologic entities including gastrointestinal malignancies, gynecologic malignancies, etc.

The strength of our study lies in inclusion of 4 lymphoid diseases and in focusing on the more frequent NHL types, taking DP a step closer to practical pathology work. Moreover, we included in this project 128 cases collected from two databases generated at different institutions. This variety of cases from different populations and institutions combined with the successful results confirmed that our algorithm surpasses the inter-laboratory variations in the tissue processing as well as the quality and type of slides staining. This contrasts with the human eyes that must adapt to any modification of the staining, a difficult and time-consuming process. On the other hand, the current limitations of our preliminary study consist first in including only four histologic categories, not yet practical for actual clinical use in hematopathologic diagnosis. Second, the WSI collections of the two databases were obtained using the same instrument platforms and we have not attempted to use the complex stain color normalization techniques to alleviate the color variations in the tissue between various staining techniques and whole slide scanners [45]. This limits extrapolation of the results to other platforms. It is important that our results are to be confirmed in future multi-center studies using different WSI instruments. The number of cases included in our study is 128, a substantial number that generates 2,560 digital images but may still be considered limited for a DL projects which typically include many more [33]. Since DL performs better with a large sample volume, we could artificially increase our database in the future by applying the "Data Augmentation" methods such as random cropping, image rotation, image inversion, etc. [46]. Finally, future design of CNN model could benefit from a process known as "transfer learning" that helps improve the training method. Transfer learning is based on exploiting a pre-trained algorithm and calibrating it for our application. The rationale behind applying the technique resides in the fact that a pre-trained network (such as one for gynecology or gastroenterology) has already learned to extract abstract features from the images and this network can be expanded to hematopathology; a process that will speed up training the model [47-48].

**CONCLUSION**

In summary, our preliminary study provided a proof of concept for incorporating automated lymphoma diagnostic screen using digital microscopic images into the pathology workflow to augment the pathologists' productivity. Future studies will need to include far more histologic entities and many more cases for training, validation, and testing. Once this has been achieved, the CNN model is potentially suitable to improve the efficiency of the diagnostic process in histopathology. This could in turn lead to adapted protocols, where pathologists perform a more thorough analysis on difficult cases, as the straight-forward cases have already been handled by a DL system. Most researchers believe that within next 15 years, DL-based applications will play an essential role in the pathology laboratory, working alongside with pathologists to provide a more timely and accurate diagnosis.